\documentclass[runningheads]{llncs}
\usepackage{graphicx}
\usepackage{amsmath,amssymb} 
\usepackage{rotating}
\usepackage{multirow}

\usepackage{color}
\usepackage[graphicx]{realboxes}
\usepackage{cite}
\usepackage{multicol}
\usepackage[width=122mm,left=12mm,paperwidth=146mm,height=193mm,top=12mm,paperheight=217mm]{geometry}
\usepackage{capt-of}

\DeclareMathOperator*{\argmin}{\arg\!\min}
\DeclareMathOperator*{\argmax}{\arg\!\max}

\begin{document}
\newcommand{\etal}{\mbox{\emph{et al.\ }}}
\newcommand{\ie}{\mbox{\emph{i. e.\ }}}
\newcommand{\etals}{\mbox{\emph{et al.\ }'s }}

\newcommand{\tcb}[1]{\textcolor{red}{\textbf{#1}}}	

\renewcommand\floatpagefraction{.9}
\renewcommand\topfraction{.9}
\renewcommand\bottomfraction{.9}
\renewcommand\textfraction{.1} 
\pagestyle{headings}
\mainmatter

\title{Object Detection, Tracking, and Motion Segmentation for Object-level Video Segmentation} 

\titlerunning{Object-level Video Segmentation}

\authorrunning{Benjamin Drayer and Thomas Brox}

\author{Benjamin Drayer and Thomas Brox}

\institute{Department of Computer Science, \\ University of Freiburg, Germany \\ \texttt{\{drayer, brox\}@cs.uni-freiburg.de}}

\maketitle

\begin{abstract}
We present an approach for object segmentation in videos that combines frame-level object detection with concepts from object tracking and motion segmentation. 
The approach extracts temporally consistent object tubes based on an off-the-shelf detector. 
Besides the class label for each tube, this provides a location prior that is independent of motion.
For the final video segmentation, we combine this information with motion cues. 
The method overcomes the typical problems of weakly supervised/unsupervised video segmentation, such as scenes with no motion, dominant camera motion, and objects that move as a unit. 
In contrast to most tracking methods, it provides an accurate, temporally consistent segmentation of each object. 
We report results on four video segmentation datasets: YouTube Objects, SegTrackv2, egoMotion, and FBMS. 

\end{abstract}
%
%
\section{Introduction}
Video object segmentation plays a role in many high level computer vision tasks, such as action and event recognition. 
In contrast to single images, videos provide motion as a very strong bottom-up cue that can be exploited to support the high level tasks.  

For this reason, video segmentation is often approached with unsupervised, purely bottom-up methods \cite{Lee2011, Li2013, FVS, Dong14, Ochs14, Keuper2015, Yang2015, Wang2016, yang2016}. 
Especially motion segmentation can work quite well in a bottom-up fashion, if the objects of interest show some independent motion in the video. However, this is not the case in all videos.
Very often, objects of interest are mostly static and almost all motion is due to camera motion. In such cases, motion segmentation fails. 
Also in cases where objects are moving jointly, such as a horse and its rider, a separation of the objects is often not possible with just bottom-up cues. 

These limitations are avoided by adding user input that decides in these cases~\cite{Vijayanarasimhan2012,Badrinarayanan2013,Jain2014,Naveen}.
However, this is not an option for a system that is supposed to automatically interpret video material. 

In this paper, we propose a weakly supervised method. The weak supervision
is due to the use of an off-the-shelf detector which was trained in a supervised manner on annotated images. However, running the video segmentation on new videos does not require any user input anymore. 
Our technical contribution is an effective way to combine the concept of tracking-by-detection with concepts from motion segmentation and local appearance cues from the object detector.
Typically failure cases in video segmentation, such as constant motion, jointly moving objects, as well as objects that move into the field of view, are well handled.
We provide ablation studies for the proposed tube extraction and the segmentation, as well as a detailed runtime analysis.

We report results on four common video segmentation datasets: YouTube Objects \cite{Jain2014}, SegTrackv2 \cite{Li2013}, egoMotion \cite{Naveen}, and FBMS \cite{Ochs14}. 
Regarding the largest and, thus, most relevant YouTube Objects dataset, we perform $3\%$ better than the current state of the art on video segmentation methods without user interaction. 
The robustness of our method is further demonstrated by the good results on the other datasets with average performance gains of up to $16\%$. 

%
%
%
\begin{figure*}[t]
\begin{center}
   \includegraphics[width=1.0\linewidth]{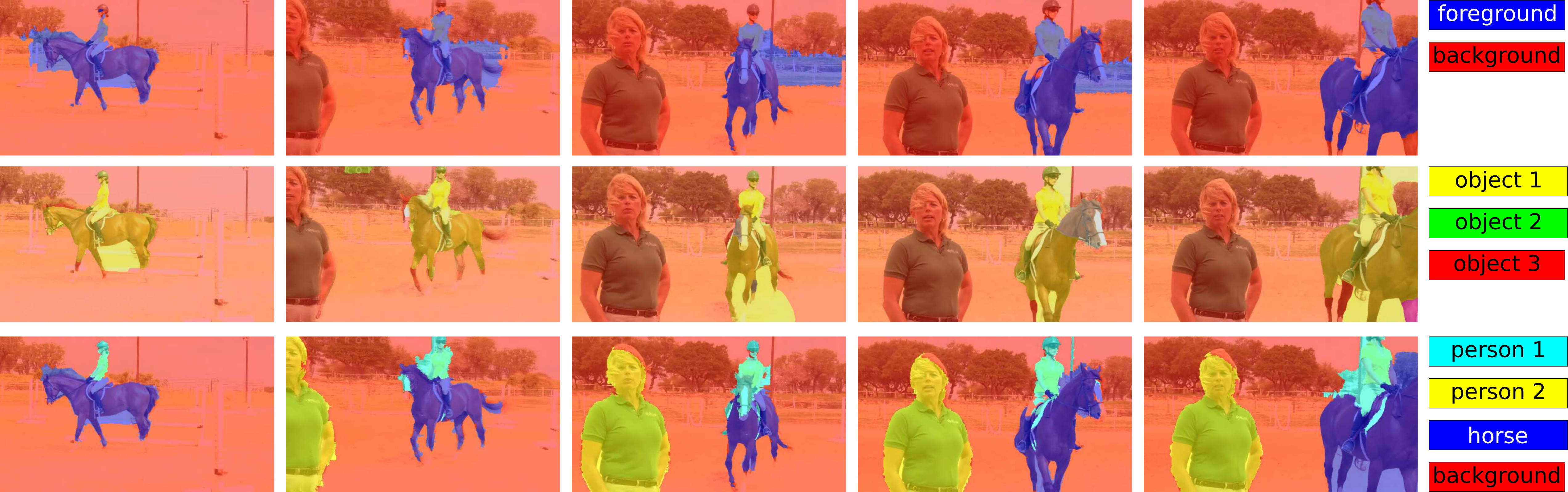}\vspace{-5mm}
\end{center}
   \caption{This example from the YouTube Objects dataset highlights  common challenges in video segmentation:
   strong camera motion, multiple object instances, and appearance of a new object. 
   Unsupervised methods as \cite{FVS} (\textbf{top row}) and \cite{Keuper2015} (\textbf{middle row}),
   fail to recognize the rider as an individual object as well as the static person on the left. Our weakly supervised method
    (\textbf{bottom row}) deals well with these issues and correctly identifies these objects. Besides the segmentation, we also 
    retrieve the class label of each object.
   }
\label{fig:teaser}
\end{figure*}
%
%
%

\section{Related Work}
Video segmentation has quite some overlap with tracking, especially in the case here, where object detections are propagated over time. 
The typical tracking scenario is based on bounding boxes and usually does not provide accurate object contours. 
Tracking-by-detection is a popular approach to find consistent tracks with only little 
supervision \cite{Andriluka08, Breitenstein09, Kalal12}. Since this is a field of its own, we only review the most related works in the context of video segmentation. 
Prest \etal \cite{Prest13} generate detections in each frame, which are subsequently tracked in a greedy fashion. 
Dong \etal \cite{Dong14} use the appearance and motion based proposals from Lee \etal \cite{Lee2011} to build a graph and extract the longest tube.
In the recent work of Weinzaepfel \etal \cite{Weinzaepfel2015}, convolutional neuronal networks
generate the features for tracking. 
The work of Hua \etal \cite{Hua14} uses some intermediate motion segmentation to model occlusion in bounding box tracking. 

Supervised video segmentation methods achieve good results at the expense of user interaction for each video to be segmented \cite{Naveen, Jain2014, Badrinarayanan2013, Vijayanarasimhan2012}. 
The most popular procedure here is to annotate a single frame and the algorithm propagates the information and segments accordingly. 
In general this works well, but as new objects enter the scene, theses methods fail or additional user input is required.

Unsupervised video segmentation is usually based on motion to a certain degree. In motion segmentation, motion is the only feature for localizing objects. 
Ochs \etal \cite{Ochs14} and Keuper \cite{Keuper2015} cluster long term point trajectories. 
Papazoglou and Ferrari \cite{FVS} use optical flow to compute  so-called inside-outside maps, partitioning the frames into foreground and background. 
Yang \etal \cite{Yang2015} use motion to detect disoccluded areas and assign them to the correct object. 
The common drawback of pure motion segmentation is the need for distinct motion of the objects and the background.
Lee \etal \cite{Lee2011} employ object proposals \cite{Endres10} to enhance motion cues with a set of static features.
A sequence of min-cuts generates the figure ground segments in the work of Li \etal \cite{Li2013}.
Multiple paths connecting the segments are extracted and post-processed, resulting in a set of multiple possible segmentations.
Dong \etal \cite{Dong14} enforce the temporal consistency of object proposals via optical flow.
Wang\&Wang \cite{Wang2016} discover reoccurring objects in the video, from which they estimate
a holistic model. Yang \etal \cite{yang2016} estimate the appearance and the segmentation simultaneously by adding
auxiliary nodes to the Markov random field model.

The work of Prest \etal \cite{Prest12} uses point trajectories like Ochs \etal \cite{Ochs14} and Keuper \cite{Keuper2015} to identify objects. 
To assign class labels to the object regions, they jointly optimize over videos with the same class label. 
Hartmann \etal \cite{Hartmann2012} and Tang \etal \cite{Tang13} also use the video tag to train a classifier for frame wise segments.

In Zhang \etal \cite{Zhang15}, frame-by-frame detections and segmentation proposals are combined to a temporally consistent semantic segmentation. 
The combination of a detector and a video segmentation approach is similar in spirit to our work, but technically, the approach is very different.
We directly compare to Zhang \etal on the YouTube dataset. 
Also in the recent work of Seguin \etal \cite{seguin2016}, object tracking (either manual or via detection) guides a multiple instance segmentation. 
However, they do not make use of motion information, thus ignoring a powerful bottom-up cue in videos. 

%
%
\section{Video Object Segmentation}
\begin{figure}[t]
\centering
   \includegraphics[width=1.0\linewidth]{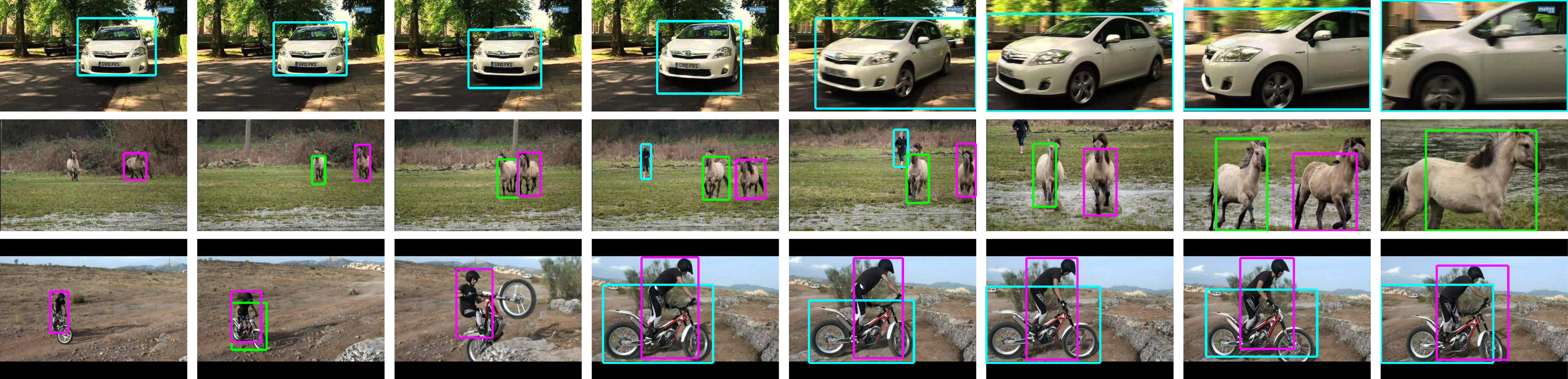}\vspace{-2mm}
   \caption{Tube extraction on three shots from the Youtube dataset. 
   Neither strong changes in viewpoint, as in the first example, nor multiple instances of the same object class, as in the second example, are a problem for our approach. 
   The third example shows a wrong classification of the motorcycle as a bicycle in the second image (green box). The method recovers later from this failure case.}
\label{fig:tube}
\end{figure}
The key input to our video object segmentation are so-called tubes. 
Individual tubes are generated by tracking the detections of an off-the-shelf R-CNN detector \cite{rcnn}. 
The subsequent spatio-temporal segmentation is guided by this initial localization and the corresponding appearance cues.

\subsection{Tube Extraction}
Good object tubes relief us from misleading motion cues, which typically occur
when the camera motion dominates the object motion, the object does not move at all, or multiple objects move as a unit.

The initial set of detections is generated by classifying the fast edge boxes from Zitnick and Dollar \cite{Zitnick14} with the R-CNN from Girshick \etal \cite{rcnn}. 
We denote the set of detected boxes with $\mathcal{B}$ and the $i$th detection in frame $t$ with $B_t^i$.
Extracting a consistent tube over time translates to finding the longest path in a graph that connects all bounding boxes of a frame with all bounding boxes in successive frames:
\begin{equation}
P^*=\argmax_{P\subseteq \mathcal{B}} \sum_{t_1<t_2} S(B_{t_1}^{i},B_{t_2}^j),
\end{equation}
where $S(\cdot,\cdot)$ measures the similarity between two detection $B_{t_1}^i$ and $B_{t_2}^j$; see Figure \ref{fig:graph} for an illustration of the graph. 
Note that there can be multiple frames between $t_1$ and $t_2$.
Consistent detections with only little or no change in appearance, position and shape reflect in high similarity values.
The exact formulation of the proposed similarity metric involves several terms, for which we refer to Section 
\ref{seq:tube_details}, where we also provide associated ablation studies.
An exemplary tube extraction is shown in Figure \ref{fig:graph} and some qualitative results are shown in Figure \ref{fig:tube}.

In general this problem is NP-hard, but in this setting where we have a directed acyclic graph, we find the global optimum in linear time using a dynamic programming approach comparable to \cite{Dong14}. 
The used topology in \cite{Dong14} might look similar to our graph, but we encode the node weights already in the edges and the similarity metric is a different one.

Apart from increasing robustness, the detector labels the objects, which can be crucial
for many high-level vision tasks.

%
%
\begin{figure}[thb]
\centering  
\includegraphics[width=1.0\linewidth]{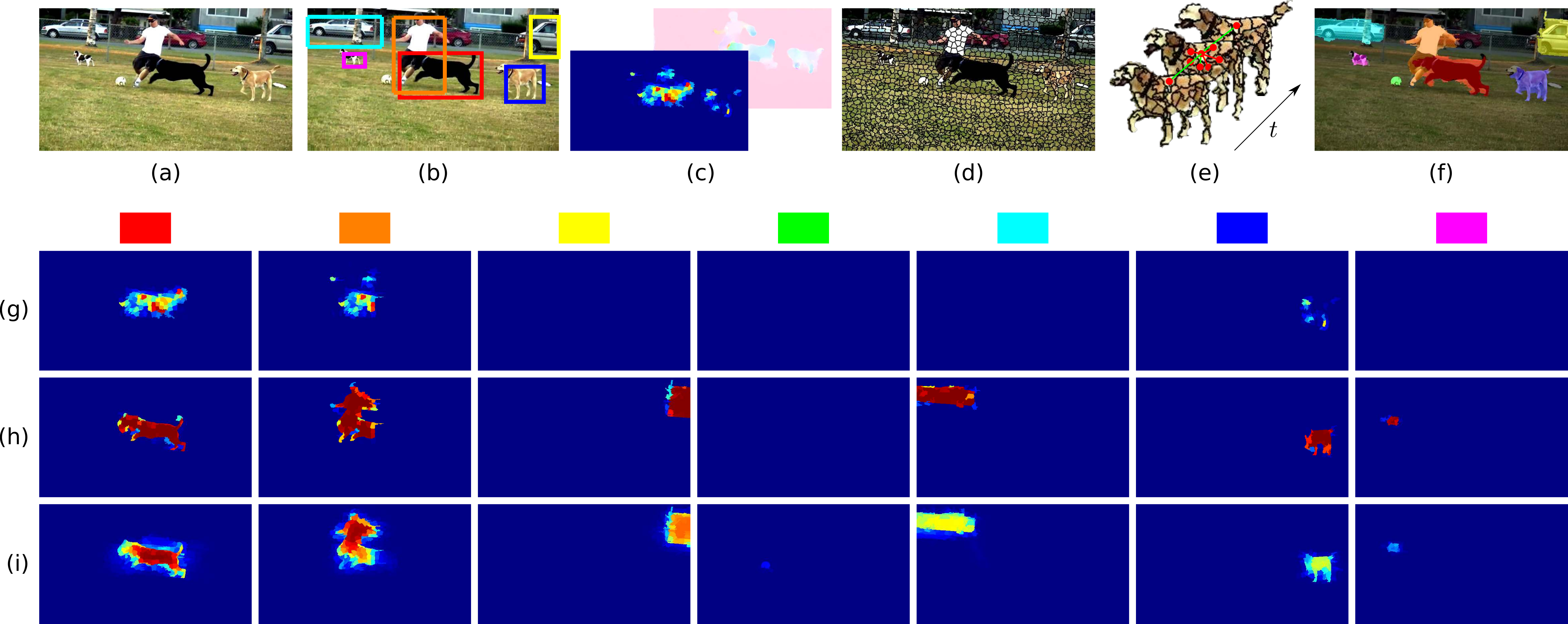}
   \caption{\textbf{Top row:} Overview of the segmentation pipeline. \textbf{(a)} One frame of the video. \textbf{(b)} In the video, we detect spatio-temporal object tubes. 
   \textbf{(c)} The optical flow and the tubes together yield inside-outside maps. \textbf{(d)} Superpixels. \textbf{(e)} Spatio-temporal graph based on these superpixels. \textbf{(f)} Final segmentation.
   \textbf{Bottom row:} Location priors for the objects detected in this example (columns correspond to the bounding boxes in the respective color). \textbf{(g)} Inside-outside maps based on optical flow. 
   \textbf{(h)} Grab cut segmentations within the detected bounding boxes. \textbf{(i)} Combination of both and propagated over time.}
\label{fig:segmentation}
\end{figure}
\subsection{Segmentation}
%
We formulate the segmentation as an energy minimization problem with a unary and a pairwise term:
\begin{equation}
 \bold{u}^*=\argmin_\bold{u} E_U(\bold{u}) + E_P(\bold{u}),
\end{equation}
where $\bold{u}$ assigns to each node in the video a label $\left\{1,\dots,K\right\}$.
The scaffold of the most supervised (e.g. \cite{Jain2014, Naveen}) and unsupervised (e.g. \cite{yang2016, FVS}) 
video segmentation algorithm is a spatio-temporal graph $G=(\mathcal{V},\mathcal{E})$ that primary enforces consistency within the frame and over time.
Long and higher resolution video shots make pixel-level segmentation computational too demanding.
Therefore the nodes $v\in\mathcal{V}$ and the final segmentation are on super-pixel level. We use superpixels from Achanta \etal~\cite{Achanta2012}.

Regarding the edges, we distinguish between spatial and temporal connections. 
Adjacent super-pixels build a spatial edge and super-pixels connected by optical flow 
build a temporal edge. For an example see Figure \ref{fig:segmentation} (e).

Similar to \cite{FVS, yang2016}, the weighting $\lambda_{(v_1, v_2)}$ of the spatial edges is proportional to the color similarity, whereas it depends on the number of matched pixels for the temporal domain.

\subsubsection{Unary Potential}
We extract the location prior $L$ and the appearance model $A$ from motion features and the tubes.
The sum builds the unary term:
\begin{equation}
 E_{U}(\bold{u})=\sum_{v\in\mathcal{V}} \left( L_{t(v)}^{\bold{u}(v)}(v) + A_{t(v)}^{\bold{u}(v)}(v) \right),
\end{equation}
where $t(v)$ is the time of super-pixel $v$ and $\bold{u}(v)$ the corresponding label.

For the location prior $L$, we first partition the inside-outside maps~$M_t$~\cite{FVS}.
These maps classify a pixel as object, if it lies within an area surrounded by motion boundaries.
We use the object tubes to restrict $M_t$:
\begin{equation}
M_t^i=B_t^i \cap M_t,
\end{equation}
with the box $B_t^i$, so that we can reliably distinguish the motion between different objects.
Additionally, motion of the background or the camera is suppressed.

We complement the motion features with foreground features $F_t^i$ by segmenting the individual boxes of the tubes with GrabCuts \cite{Rother04}. 
This feature becomes valuable, especially when the motion is constant or unreliable.

From the union of the two sets $F_t^i\cup M_t^i$, we directly compute the respective appearance models $A^i$ as Gaussian mixture models, where the background is modeled as complement of all tubes.

Temporal smoothing of the combined motion and foreground features gives us the location prior $L$.
We use an optical flow propagation in a similar fashion as \cite{FVS} to remove single bad motion or foreground estimates and carry the information beyond the endings of the tube.

A comprehensive example of this process, including the different location and foreground maps as well as the location prior $L$ is given in Figure \ref{fig:segmentation}.

\subsubsection{Pairwise Potential}
The pairwise term, enforcing spatial and temporal smoothness is a weighted Potts model:
\begin{equation}
 E_p(\bold{u})=\sum_{(v_1, v_2)\in\mathcal{E}} \delta(\bold{u}(v_1), \bold{u}(v_2)) \cdot \lambda_{(v_1, v_2)},
\end{equation}
where $\delta$ is the Kronecker delta and $\lambda_{(v_1, v_2)}$ is the edge weight.

We efficiently minimize the submodular energy with the Fast\_PD solver from Komodakis and Tziritas \cite{Komodakis2007}. 

%
%
\section{Implementation Details}
\subsection{Similarity Measure}
\label{seq:tube_details}
\begin{figure}[thb]
\centering
\includegraphics[width=1.0\linewidth]{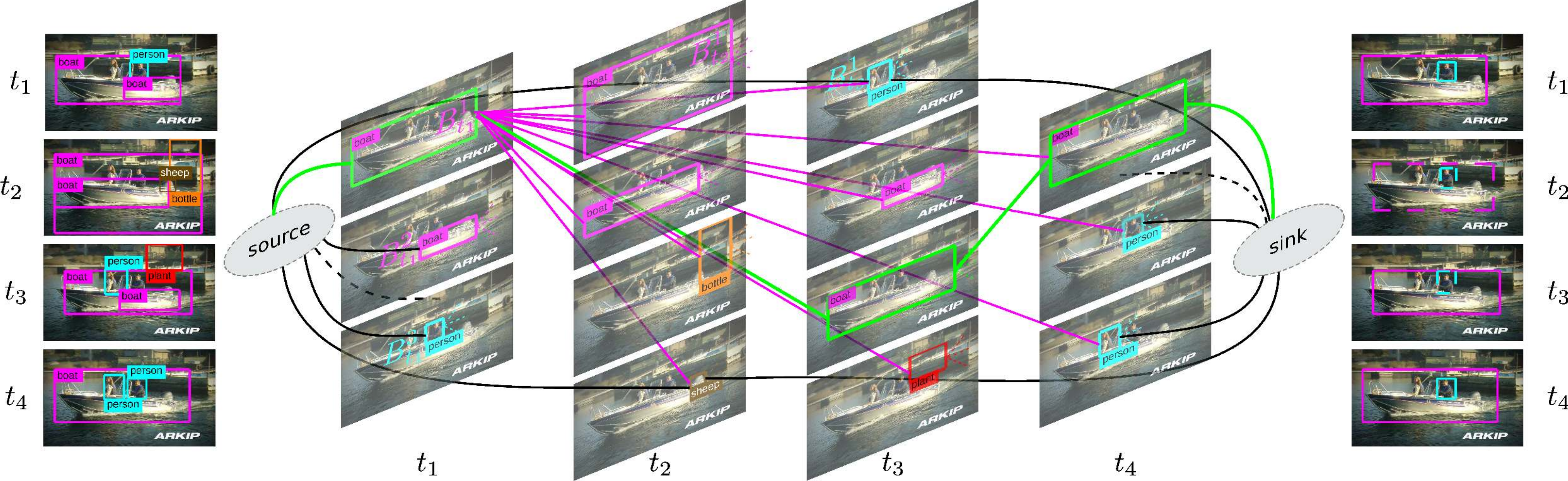}
   \caption{Illustration of the graph structure for a sample video. The left column shows all initial detections, the right column the two final high scoring tubes that have been extracted: 
   the boat (magenta) and the person (cyan). Dashed boxes indicate parts of the tube that have been interpolated between frames. 
   The graph structure is shown in the middle, where we show exemplified all edges for $B_{t_1}^1$. The corresponding longest path is shown in green.}
\label{fig:graph}
\end{figure}

Favoring consistency in position, shape and appearance, reflects in the following similarity measure
for two detections $B_{t_1},B_{t_2}$:
\begin{equation}
  \begin{split}
 S(B_{t_1},B_{t_2})=&score(B_{t_2}) \cdot S_{category}(B_{t_1},B_{t_2}) \cdot S_{app}(B_{t_1},B_{t_2}) \\
 & {\cdot S_{vol}(B_{t_1},B_{t_2}) } \cdot S_{side}(B_{t_1},B_{t_2}) \cdot S_{match}(B_{t_1},B_{t_2})\\
  &\cdot S_{center}(B_{t_1},B_{t_2}).   
  \end{split}
\end{equation}
With the category label, the appearance and center term, we favor a consistent appearance of the object. The side and volume constraints enforce the tube to change  its shape smoothly.
Temporal consistency is encoded in the matching term, and the score rewards confident detections. 

The category label is a very powerful indicator and has to be consistent through time. Therefore we set 
\begin{equation}
 S_{category} = 
\begin{cases} 1 & \mbox{if } category(B_{t_1})=category(B_{t_2}) \\ 
-\infty & \mbox{else } . 
\end{cases}
\end{equation}
Due to movement of the camera and/or the object itself, the bounding box can change over time. 
This is supposed to be a rather slow process, therefore we favor small changes in both, the volume and the sides.
\begin{equation}
 S_{vol} = \min\left(\frac{Vol(B_{t_1})}{Vol(B_{t_2})},\frac{Vol(B_{t_2})}{Vol(B_{t_1})}\right) 
\end{equation}
The cost for the side change is computed in the same way, where we take the minimum of the height and the width change. 
The matching term gives the ratio on how many points of $B_{t_2}$ are matched by the optical flow $F$ originating from $B_{t_1}$.  
\begin{equation}
  \begin{split}
 S_{match} &= \frac{\left|Matches\right|}{Vol(B_{t_2})} \\
 Matches&=\left\{ \bold{p}\in B_{t_2}\vert\exists \bold{q}\in B_{t_1} : F(\bold{q})=\bold{p}\right\}
 \end{split}
\end{equation}
Although the optical flow is an indicator on how similar the two boxes are, the volume of $B_{t_2}$ and 
the possible distinct motion of object and background weaken this term. 

We compensate for that by additionally penalizing the deviation of the propagated center $c_p$ of box
$B_{t_1}$ with the actual center $c$ of $B_{t_2}$. 
\begin{equation}
 S_{center} = \frac{1}{1 + 0.1 \cdot \left\| \boldmath{c}_p- \boldmath{c} \right\| }
\end{equation}
Correlating $B_{t_1}$ with frame $t_2$ gives us the propagated center $c_p$. 
On a finer level, this is less accurate than optical flow, but it is more robust.

The appearance term is the cosine-distance of color histograms $H(\cdot)$:
\begin{equation}
 S_{app} = \frac{\langle H(B_{t_1}) , H(B_{t_2}) \rangle}{ \left\| H(B_{t_1}) \right\| \cdot \left\| H(B_{t_2}) \right\|}.
\end{equation}
This cue is independent of the area and shape given by the corresponding bounding box.
Boxes with $S_{app}\leq0.8$ are considered as distinct and the term is set to $-\infty$. When objects of the same class interact (e.g. overtaking cars, Figure \ref{fig:tube_components}),
the appearance is important to track them correctly. 

\subsection{Graph}
We build the graph by connecting the detections in a temporal order. Since we cannot assume that the detections are present in every frame we interconnect each detection with the detections 
of the subsequent $20$ frames.

Additionally, each node is connected to the source and sink, so that new objects can enter and leave the scene, while being correctly tracked without introducing additional knowledge about 
the object's presence in the video. 
See Figure \ref{fig:graph} for a visualization of such a graph.

\subsection{Post-processing}
We interpolate the possible sparse tube into a dense one. 
The missing box in frame $t$ is interpolated by correlating the box found in frame $(t-1)$ with frame $t$. 
Gaps of more than one frame are interpolated from both sides.

The set of tubes is cleaned by a volumetric non-maximum suppression, where overlapping tubes of the same class with an intersection over union $>0.5$ are suppressed by the longer, respective higher scoring tube.

\subsection{Tube Parameter Evaluation}
\begin{figure}[t]
\centering
   \includegraphics[width=1.0\linewidth]{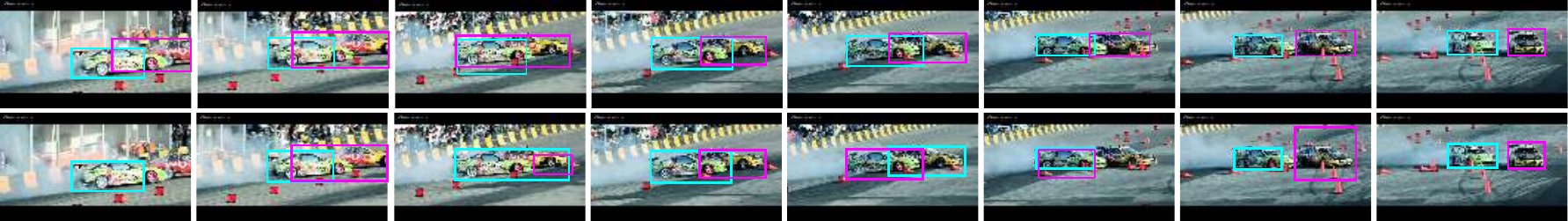}
   \caption{Relevance of the appearance constraint $S_{app}$. Our tube extraction result (\textbf{top row}) gets significantly worse (\textbf{bottom row}) when we drop $S_{app}$, as the two cars get mixed up.}
\label{fig:tube_components}
\end{figure}

We give a justification of the different terms in our proposed similarity measure in form of an ablation study in Table \ref{tab:tube_components}.
For the evaluation, we selected a subset of the SegTrackv2 dataset (birds\_of\_paradise, bmx, drift, girl, monkey, penguin and soldier) that excludes the categories, 
for which we have either no detector or not sufficient detections. With no or little detections, tracking is almost independent of different similarity measures. 
Note that we favored the SegTrackv2 dataset as it provides annotation for every frame and allows for a more detailed analysis. Having only every 10th frame annotated (YouTube), 
flickering or swapping of labels between objects (as in the drift sequence, Figure \ref{fig:tube_components}) will be missed by the evaluation metric.

We report the quality of the tube as IoU (intersection over union) between the boxes of the tube
and the boxes spanned by the GT-annotation. 
We judge the impact of the different components by two experiments, first we drop the component (-)
and measure how the performance decreases and second we use the specific component as sole similarity measure (+).
While the best performance is achieved by using all components, the performance depends most on the appearance term,
which is exemplified in Figure \ref{fig:tube_components}.

%
%

\begin{table}
\begin{center} 
\begin{tabular}{|c|c|c|c|c|c|c|c|c|c|}
\hline
                  &   $score$   & $S_{side}$    & $S_{vol}$  & $S_{vol}$ $S_{side}$   & $S_{match}$ & $S_{center}$  & $S_{match}$ $S_{center}$& $S_{app}$ & all \\
\hline
           -  &   $58.75$   & $59.7$        & $59.5$     & $59.0$                 & $57.8$      & $59.13$         & $57.85$                   & $55.73$   & $\bf{59.83}$\\
\hline
             +  &   $53.47$   & $55.07$       & $54.33$    & $53.99$                & $54.75$     & $54.81$         & $54.21$                   & $57.39$   & $\bf{59.83}$\\
\hline
 \end{tabular}
\end{center}
\caption{Ablation study of the tube extraction on a subset of SegTrackv2, reported as IoU. Removing components from the system makes results worse (-). 
Some components are more important than others. The performance of the individual components (+) confirm that the appearance term is the most significant}
\label{tab:tube_components}
\end{table}

%
%
\subsection{Resolving Oversegmentation}
When objects suddenly deform or rapidly change their appearance, it is likely that they are tracked by multiple tubes.
Lowering the restriction of the tube is not a solution since the tubes would lose consistency. Especially when
tracking multiple objects, it is important that the tube does not jump between different objects (see Figure \ref{fig:tube_components}).

\begin{figure}
\centering
   \includegraphics[width=1.0\linewidth]{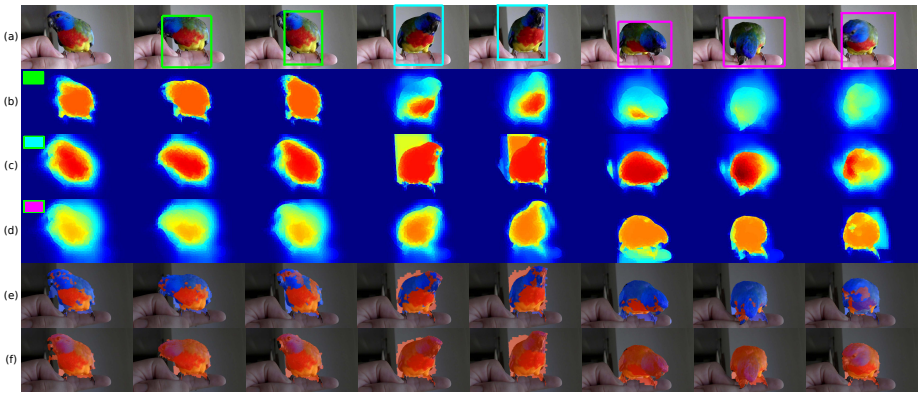}
   \caption{Multiple tubes tracking the same object \textbf{(a)} are a result of quick changes in appearance, shape or position and
   lead to oversegmentation \textbf{(e)}. Using the location cues \textbf{(b-d)}, we merge the tubes and consistently segment the 
   bird \textbf{(f)}.}
\label{fig:mergestep}
\end{figure}

Multiple tubes lead to a later oversegmentation. We use the location prior $L$ and the class label, to merge
cohesive tubes. If the correlation between different locations priors with the same class label is $\geq 0.5$,
we fuse the tubes. Figure \ref{fig:mergestep} gives an typical example.

%
%

\section{Results}
With the evaluation on four video segmentation datasets (YouTube Objects \cite{Jain2014}, egoMotion \cite{Naveen},
SegTrackv2 \cite{Li2013} and FBMS \cite{Ochs14}), we prove the performance of the proposed method.
The datasets impose different challenges and shortcomings.

EgoMotion and FBMS are complementary. In egoMotion, there is always a single object that is largely static,
whereas FBMS contains multiple moving objects. The downside of FBMS is that there is no ground-truth annotation for static objects, because it is a benchmark dataset designed for motion segmentation.  

SegTrackv2 and YouTube Objects are the most relevant datasets, since they are composed of a variety of different settings.
For video level segmentation YouTube Objects have become the dataset on which state-of-the-art methods report their results. 
%
%
\subsection{YouTube}
\begin{figure}
\centering
   \includegraphics[width=1.0\linewidth]{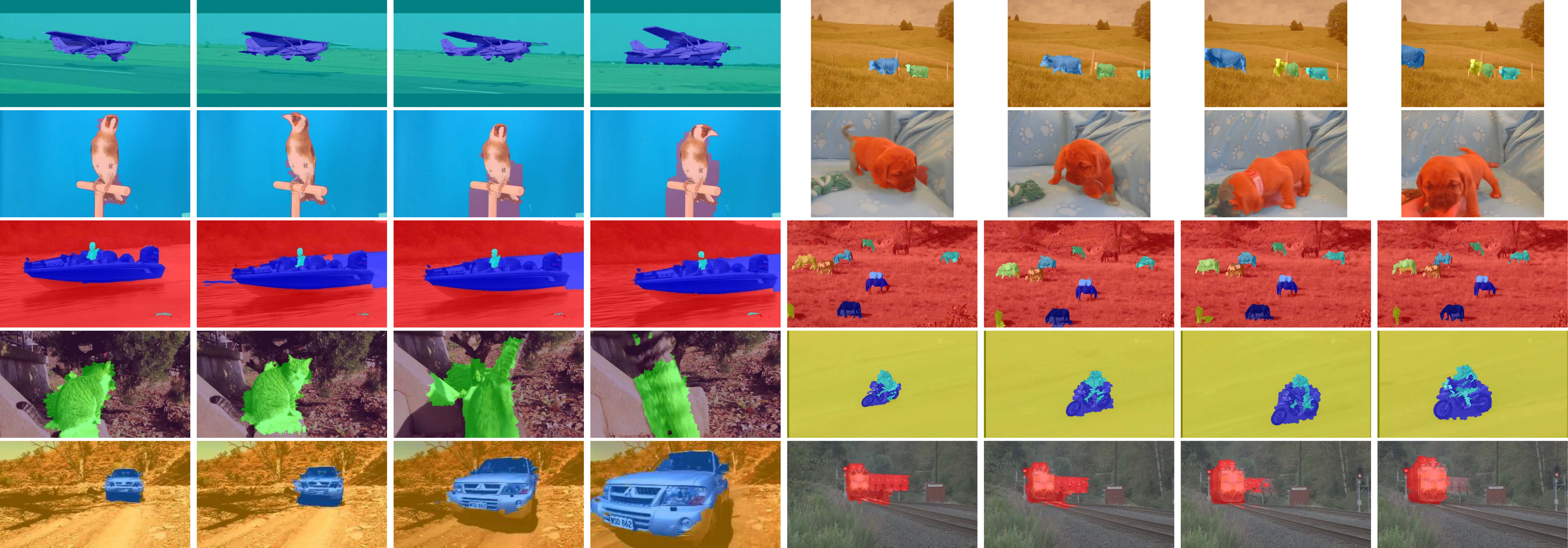}
   \caption{Results for the 10 different object categories of the YouTube dataset. The proposed method distinguishes connected objects (boat, motorbike). Rapidly moving, non-rigid objects (cat) cause some problems.}
\label{fig:youtube}
\end{figure}
126 Videos from 10 different classes make up the YouTube Objects dataset \cite{Prest12}.
Jain \etal \cite{Jain2014} provide ground-truth annotation on a super-pixel level for approximately every 10th frame.
The evaluation metric is the average intersection over union 
\begin{equation}
IOU=\frac{\left| S\cap GT \right|}{\left| S\cup GT \right|},
\end{equation}
where $S$ is the segmentation and $GT$ the ground truth.

With the proposed tubes, an implicit segmentation is already given by the foreground features $F$ (tube cut).
Besides our full segmentation, we evaluate two further approaches.
First we use only the partitioned motion features $M_t^i$ as prior for the segmentation (OURS-M).
Analogously, we evaluate the segmentation based on pure appearance features $F_t^i$ (OURS-F).

The results in Table \ref{tab:youtube} show that the foreground features without further processing 
perform similar to pure motion based segmentation. The best result is achieved by the combination
of motion and appearance cues, which beats current state of the art methods by at least $3\%$.

\begin{table}
 \begin{center}
  \begin{tabular}{|c|c|c|c|c|c|c|c|c|c|c||c|c|}
  \hline
                   & aero     & bird   & boat   & car    & cat    & cow    & dog    & horse  & motorbike & train  & avg cat & avg vid\\
  \hline
     \cite{Ochs14} &  $13.7$  & $12.2$ & $10.8$ & $23.7$ & $18.6$ & $16.3$ & $18.0$ & $11.5$ & $10.6$    & $19.6$ & $15.5$  & $-$\\
  \hline
     \cite{Tang13} &  $17.8$  & $19.8$ & $22.5$ & $38.3$ & $23.6$ & $26.8$ & $23.7$ & $14.0$ & $12.5$    & $40.4$ & $23.9$  & $22.8$\\
  \hline
    \cite{FVS}   &  $65.5$  & $69.2$ & $43.9$ & $66.1$ & $49.3$ & $38.0$ & $50.0$ & $31.6$ & $31.9$    & $34.0$ &  $47.9$ & $45.7$ \\
  \hline
     \cite{Zhang15}&  $75.8$  & $60.8$ & $43.7$ & $71.1$ & $46.5$ & $54.6$ & $55.5$ & $54.9$ & $42.4$    & $35.8$ &  $54.1$ & $52.6$ \\
  \hline
    \cite{Wang2016}&  $63.0$  & $69.0$ & $40.0$ & $61.0$ & $48.0$ & $46.0$ & $67.0$ & $53.0$ & $47.0$    & $38.0$ & $53.0$  & $-$\\
  \hline
    \cite{Long2015}&  $59.3$  & $67.6$ & $32.6$ & $50.5$ & $33.1$ & $27.4$ & $35.6$ & $46.0$ & $18.4$    & $47.3$ & $41.8$  & $37.3$\\
\hline
\hline
  tube cut         &  $72.3$  & $46.4$ & $56.6$ & $47.0$ & $30.1$ & $55.7$ & $39.4$ & $47.9$ & $35.1$    & $36.7$ &
  $46.7$  & $45.8$  \\ 
\hline
  OURS-$M$         &  $65.0$  & $72.7$ & $49.1$ & $68.9$ & $49.9$ & $49.6$ & $54.4$ & $39.0$ & $37.2$    & $37.0$ &
  $52.3$  & $51.2$  \\ 
\hline
  OURS-$F$         &  $73.3$  & $67.0$ & $60.0$ & $57.3$ & $34.5$ & $62.4$ & $54.7$ & $54.6$ & $42.1$    & $38.1$ &
  $54.4$  & $54.0$  \\ 
  \hline
  OURS             &  $74.4$  & $72.1$ & $58.5$ & $60.0$ & $45.7$ & $61.2$ & $55.2$ & $56.6$ & $42.1$    & $36.7$ & $\bf{56.2}$  & $\bf{55.8}$  \\ 
\hline
  \end{tabular}  
 \end{center}
 \caption{Results for the YouTube dataset, reported as IoU. The proposed method performs $3\%$ better than the current state-of-the-art method \cite{Zhang15}. The combination of motion and appearance features lead to the best performance.}
 \label{tab:youtube}
\end{table}

%
%
\subsection{SegTrackv2}
\begin{figure}
\centering
\begin{minipage}[t]{.45\textwidth}
\centering
\vspace{0pt}
\scriptsize{
\begin{tabular}{|c|c|c|c|c|c|}
\hline
			& \cite{Ochs14} & \cite{Keuper2015} & \cite{FVS} & \cite{Long2015} & OURS\\
\hline
bird\_of\_paradise 	& $17.2$ & $\bf{79.0}$ & $74.9$ & $43.2$ & $50.5$ \\
\hline
birdfall           	& $0.5$  & $0.5$  & $\bf{4.5}$  & $-$ & $\bf{4.5}$ \\
\hline
bmx-person 		& $4.8$  & $70.4$ & $47.8$ & $0.9$ & $\bf{90.7}$ \\
\hline
bmx-bike   		& $1.2$  & $17.3$ & $16.3$ & $20.0$ & $\bf{33.5}$\\
\hline
cheetah-deer  		& $1.9$  & $1.9$ & $\bf{47.1}$ & $-$ & $41.9$\\
\hline
cheetah-cheetah		& $0.9$  & $3.9$ & $\bf{17.9}$ & $-$ & $0$\\
\hline
drift-car1		& $35.1$ & $50.2$ & $48.4$ & $36.0$ & $\bf{70.1}$\\
\hline
drift-car2		& $12.4$ & $0.3$  & $35.0$ & $39.6$ & $\bf{60.2}$\\
\hline
frog			& $41.5$ & $43.1$ & $57.3$ & $-$ & $\bf{68.4}$\\
\hline
girl			& $52.1$ & $51.4$ & $53.8$ & $\bf{65.8}$ & $65.4$\\
\hline
monkey			& $35.8$ & $22.8$ & $64.8$ & $-$ & $\bf{65.2}$\\
\hline
monkeydog-dog		& $1.4$  &  $\bf{6.8}$ & $0$    & $0$ & $0$\\
\hline
monkeydog-monkey	& $54.9$ & $54.9$ & $\bf{77.7}$ & $0$ & $\bf{77.7}$\\
\hline
hummingbird-bird1	& $3.9$  & ${11.0}$ & $10.1$ & $\bf{39.8}$ & $10.4$\\
\hline
hummingbird-bird2	& $\bf{55.4}$ & $32.4$ & $51.6$ & $30.2$ & $9.4$\\
\hline
parachute		& $\bf{90.3}$ & $89.8$ & $68.7$ & $-$ & $68.7$\\
\hline
penguin-penguin1	& $8.5$  & $8.6$ & $4.7$ & $20.0$ & $\bf{43.0}$\\
\hline
penguin-penguin2	& $3.8$  & $\bf{4.1}$ & $1.9$ & $0.7$ & $0$\\
\hline
penguin-penguin3	& $0$    & $3.8$ & $1.7$ & $\bf{14.9}$ & $0$\\
\hline
penguin-penguin4	& $0$    & $0$   & $\bf{2.2}$  & $0$ & $0$\\
\hline
penguin-penguin5	& $0$    & $0$   & $\bf{8.9}$ & $0$ & $0$\\
\hline
penguin-penguin6	& $0$    & $5.3$ & $18.3$ & $0.61$ & $\bf{73.6}$\\
\hline
soldier			& $63.0$ & $50.1$ & $36.9$ & $0$ & $\bf{64.0}$\\
\hline
worm			& $2.7$  & $23.0$ & $\bf{69.0}$ & $-$ & $\bf{69.0}$\\
\hline
\hline
avg obj			& $27.9$ & $34.5$ & $43.7$ & $24.8^*$ & $\bf{48.0}$\\
\hline
avg vid			& $20.3$ & $26.3$ & $34.2$ & $18.3^*$ & $\bf{40.3}$\\
\hline
\end{tabular}  
}
\label{tab:segtrack}
\captionof{table}{Quantitative results for the SegTrackv2 dataset, reported as IoU. 
$(*)$ Results are averaged over the videos containing objects from the $20$ pascal classes.  }
\vspace{0.5cm}
\normalsize{	
\begin{tabular}{|c|c|c|c|c|c|}
\hline
              &  car    & cat    & chair    & dog & average \\
\hline
 \cite{Ochs14}     &  $33.6$ & $13.5$ & $16.2$ & $41.7$ & $26.3$ \\
\hline
 \cite{Keuper2015} &  $37.9$ & $45.3$ & $19.8$ & $53.4$ & $39.1$ \\
\hline
 \cite{FVS}        &  $47.6$ & $56.6$ & $59.5$ & $64.2$ & $57.0$ \\
 \hline
 \cite{Long2015}   &  $86.1$ & $16.6$ & $39.0$ & $47.1$ & $47.2$ \\
\hline
 OURS              &  $\bf{78.0}$ & $\bf{65.7}$ & $\bf{73.5}$ & $\bf{75.2}$ & $\bf{73.1}$ \\
\hline
 \end{tabular}
 }
\captionof{table}{Results reported as IoU for the egoMotion dataset \cite{Naveen}. 
We have a $16\%$ improvement compared to motion segmentation approaches.}
\label{tab:ego}

\end{minipage}
\hspace{0.5cm}
\begin{minipage}[t]{.45\textwidth}
\centering
\vspace{0pt}
\includegraphics[width=\textwidth]{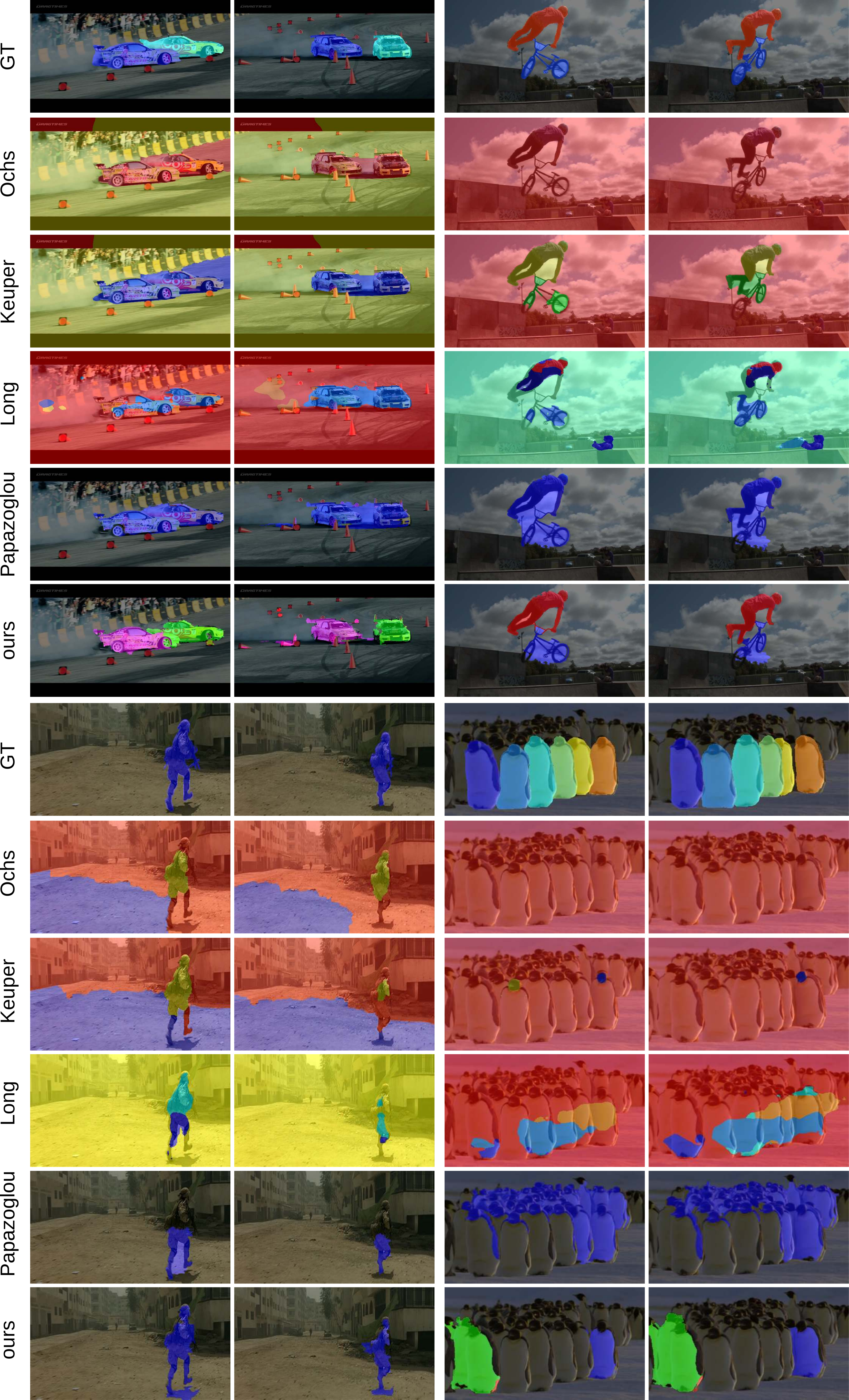}
\caption{Results for the SegTrackv2 dataset. From top to bottom: ground truth, Ochs \etal \cite{Ochs14}, Keuper \etal \cite{Keuper2015},
   Papazoglou and Ferrari \cite{FVS} and ours. The example from the drift sequence shows that motion based methods fail when the objects are close to 
   each other and move similarly. In the second frame the two pylons in the upper left corner are detected as traffic lights. For the penguin sequence, we detected only two tubes and thus did not segment the remaining penguins.}
   \label{fig:segtrack}
\end{minipage}\hfill
\end{figure}

The SegTrackv2 dataset \cite{Li2013} consists of 14 videos with frame-wise ground-truth annotation.
Single and multiple objects, slow and fast motion, as well as occluding and interacting objects are present.
Note that only in sequences, where known objects are present, our method can perform well.
When processing sequences with unknown objects such as parachute or worm, we can only rely on motion features and we fall back to the approach in \cite{FVS}.
Regarding cases, in which we can extract tubes, e.g. bmx or drift sequence, we clearly outperform the other methods.
On average, we are $4.3\%$ better than the other methods; see Table 3.
Qualitative results and comparisons are shown in Figure \ref{fig:segtrack}.

%
%
\subsection{egoMotion}
The egoMotion dataset \cite{Naveen} consists of 24 videos from 4 categories (cars, cats, chairs, dogs),
where each video features a single object.
The main challenge is the dominant camera motion, which is hard to handle for pure motion based methods. The extraction of the tubes give us a vital prior, resulting in $73.1$ (IoU),
whereas the best motion based method \cite{FVS} achieves $57.0$; compare Table \ref{tab:ego}.

\begin{figure}[t]
\centering
   \includegraphics[width=1.0\linewidth]{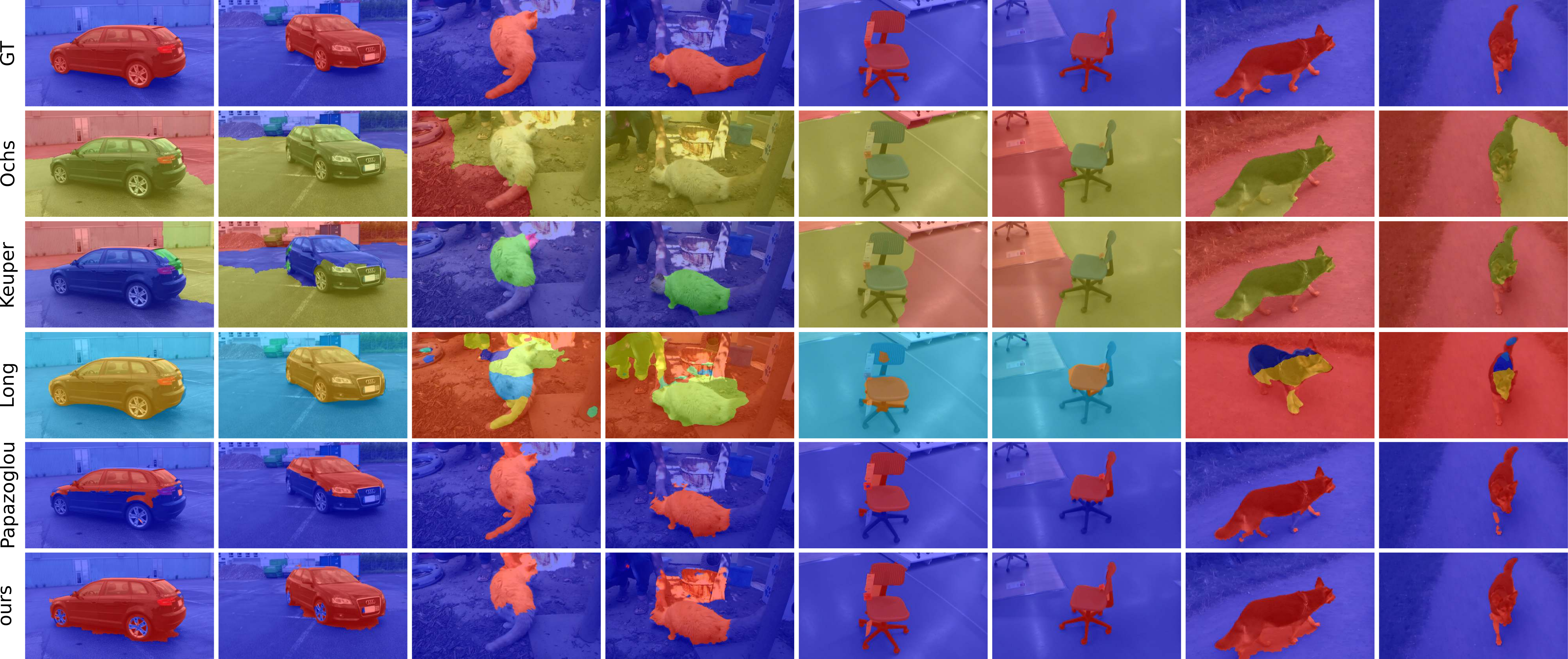}
   \caption{Results for the egoMotion dataset \cite{Naveen}. \textbf{From top to bottom:} Ground truth, Ochs \etal \cite{Ochs14}, Keuper \etal \cite{Keuper2015}, Long \etal \cite{Long2015}, Papazoglou and Ferrari \cite{FVS} and ours. For static objects, the image motion at the bottom of the object and the ground is the same. Thus, pure motion segmentation methods fail.}
\label{fig:egomotion}
\end{figure}

%
%
\subsection{FBMS}
\begin{figure}[t]
\centering
   \includegraphics[width=1.0\linewidth]{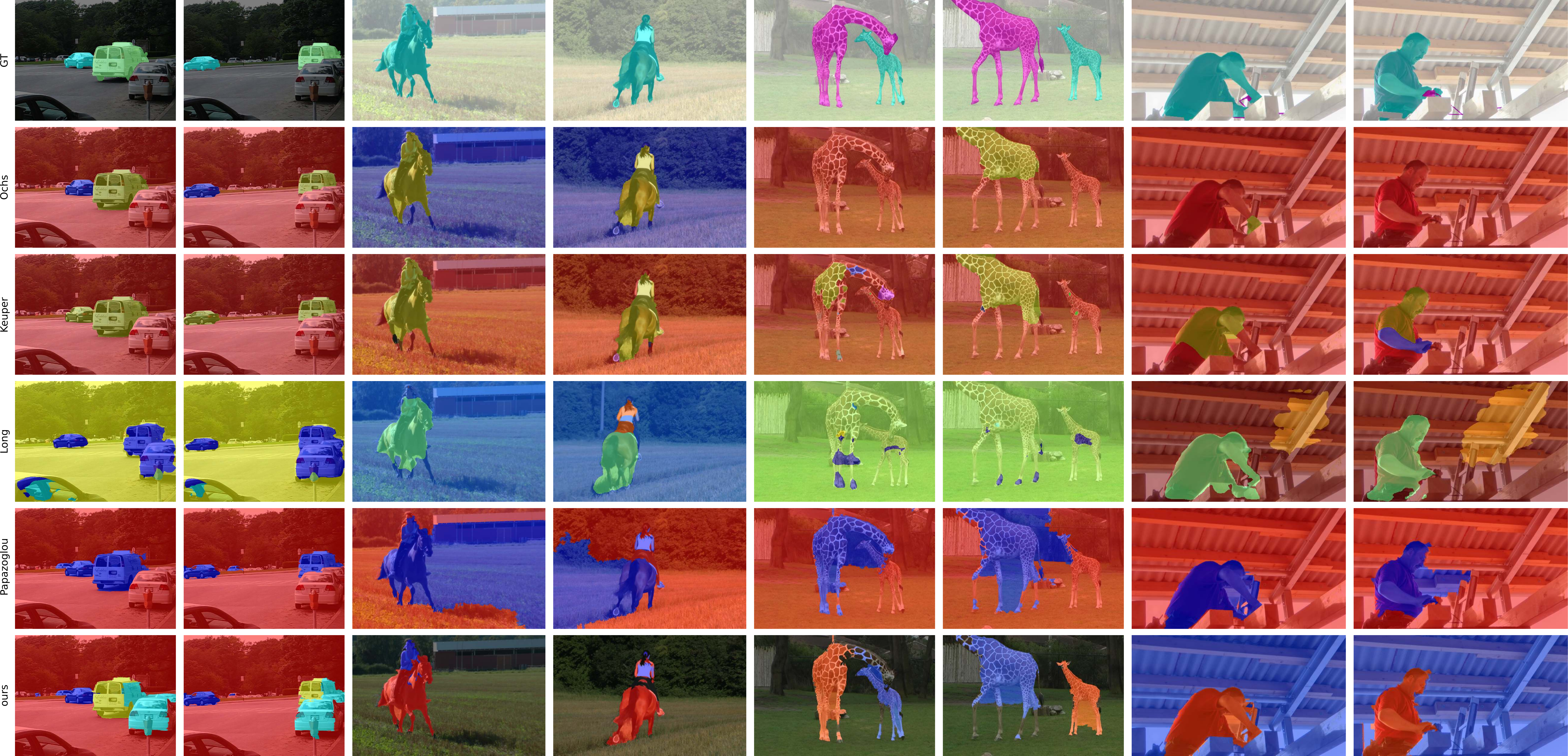}
   \caption{Qualitative results for the FBMS dataset \cite{Ochs14}. From top to bottom: ground truth, Ochs \etal \cite{Ochs14}, Keuper \etal \cite{Keuper2015},
   Long \etal \cite{Long2015}, Papazoglou and Ferrari \cite{FVS} and ours. Typical failure cases
   of our method on this dataset are: the segmentation of static objects, e.g. the white car (segmented in cyan) in the car sequence. 
   Splitting moving objects, e.g. the rider and horse. For the giraffe sequence, the extracted tubes 
   switch the object leading to an inconsistent segmentation. 
   }
\label{fig:fbms}
\end{figure}

The Freiburg-Berkely-Motion-Segmentation dataset \cite{Ochs14} shows a great variety of different moving objects over 59 videos.
Regarding the evaluation we follow \cite{Ochs14} and report the results as average precision and recall, where 
\begin{align}
 P_{i,j}=\frac{\left| S_i\cap GT_j \right|}{\left| S_i \right|}
 &&  R_{i,j}=\frac{\left| S_i\cap GT_j \right|}{\left| GT_i \right|}
 &&  F_{i,j}=\frac{2  P_{i,j}  R_{i,j}}{P_{i,j} + R_{i,j}}
\end{align}
The assignments between segmentations and ground-truth are chosen so that the F-measure is maximized. 
An object is counted as successfully segmented, if $F\geq 0.75$, where the background does not count as an object, so the number is reduced by $1$. Our method clearly performs better than \cite{Ochs14} and \cite{FVS}, both in terms of the F-measure and the
number of segmented objects. However, \cite{Keuper2015} and \cite{Yang2015} achieve better results; compare Table \ref{tab:fbms}. 
In Figure \ref{fig:fbms}, we reveal some of the cases that reduce our performance. 
The missing annotation of static objects (parking cars) and the labeling of different objects as one moving unit (rider on horse) decrease our performance. 
\begin{table}
\begin{center} 
\begin{tabular}{|c||c|c|c|c||c|c|c|c|}
\hline
	      & \multicolumn{4}{|c||}{Training} & \multicolumn{4}{|c|}{Test} \\
\hline
               &   P         & R         & F         & F $\geq75\%$ &   P         & R         & F         & F $\geq75\%$\\
\hline
\cite{Ochs14}      &  $81.50\%$ & $63.23\%$ & $71.21\%$ & $16/65$      &  $74.91\%$ & $60.14\%$ & $66.72\%$ & $20/69$\\
\hline
\cite{Keuper2015}  &  $85.31\%$ & $68.70\%$ & $76.11\%$ & $24/65$      &  $85.95\%$ & $65.07\%$ & $74.07\%$ & $23/69$\\
\hline
\cite{FVS}         &  $85.86\%$ & $61.85\%$ & $71.90\%$ & $13/65$      &  $88.72\%$ & $54.84\%$ & $67.78\%$ & $14/69$\\
\hline
\cite{Yang2015}    &  $89.53\%$ & $70.74\%$ & $\bf{79.03}\%$ & $\bf{26/65}$      & $91.47\%$ & $64.75\%$ & $\bf{75.82}\%$ & $\bf{27/69}$\\
\hline
\cite{Long2015}    &  $79.03\%$ & $63.66\%$ & $70.52\%$ & $13/65$      & $75.36\%$ & $59.66\%$ & $66.60\%$ & $19/69$\\
\hline
OURS               &  $84.27\%$ & $66.48\%$ & $74.32\%$ & $20/65$      & $86.76\%$ & $63.28\%$ & $73.18\%$ & $23/69$\\
\hline
 \end{tabular}
\end{center}
\caption{Quantitavie evaluation on the FBMS dataset \cite{Ochs14}. }
\label{tab:fbms}
\end{table}

%
%
\subsection{Runtime and Scalability}
The average runtime is $\sim 8$ seconds per frame. Table \ref{table:runtime_video_segmentation_total} gives 
an detailed overview of the contribution of the different components.
The main costs (in seconds) are caused by running the object detector $(1.53)$, GrabCuts $(2.1)$, Optical Flow $(1.04)$ and the correlation of 
boxes $(1.52)$.

\begin{figure}[t]
\centering
   \includegraphics[width=0.7\linewidth]{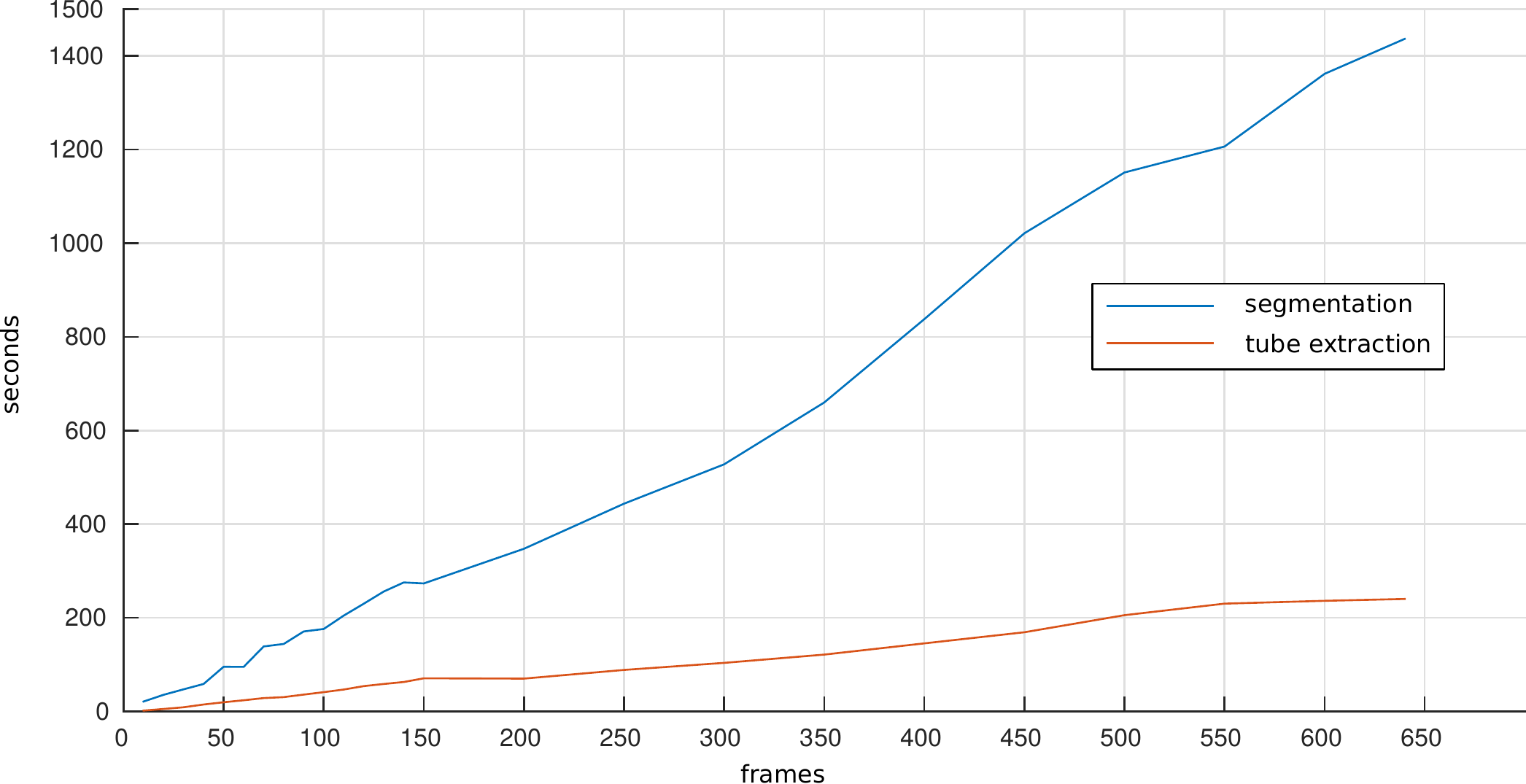}
   \caption{The runtime of the segmentation (blue) and tube (red) scales linear with the number of frames.
   Detection, optical flow and super pixel computation take constant time per frame.}
\label{fig:runtime}
\end{figure}

The scalability is analyzed in Figure \ref{fig:runtime}, where we observe a linear behavior for both, the tube extraction
and the segmentation. 

\begin{table}
  \footnotesize 
  \begin{center}
 \begin{tabular}{|c|c|c|c|c|c|c|c|c|c|c|c|c|c|c|c|}
 \hline
  \multirow{2}{*}[-2em]{\rotatebox{90}{components}}    & & \multicolumn{7}{c|}{Tube extraction} & \multicolumn{7}{c|}{Segmentation} \\
 \cline{2-16}
  & \rotatebox{90}{Optical Flow} &

  \rotatebox{90}{Detection} &  
  \rotatebox{90}{Histograms} &
  \rotatebox{90}{Correlation} &
  \rotatebox{90}{File IO} &
  \rotatebox{90}{Graph} &
  \rotatebox{90}{Longest path} &
  \rotatebox{90}{Build tube} &
  
  \rotatebox{90}{Superpixels} &
  \rotatebox{90}{File IO} &
  \rotatebox{90}{Grabcuts} &
  \rotatebox{90}{IO Maps} &
  \rotatebox{90}{Unary potentials} &
  \rotatebox{90}{Pairwise potentials} &
  \rotatebox{90}{MRF} \\
    
  \hline
  \multirow{3}{*}{\rotatebox{90}{\shortstack[l]{time\\in sec}}} &
  $1.04$ & 
  $1.53$ &  $0.58$ & $1.52$ & $0.04$ & $0.05$ & $0.004$ & $0.42$ &
  $0.34$ & $0.21$ & $2.1$ & $0.18$ & $0.28$ & $0.03$ & $0.002$\\
  \cline{2-16}
  & & \multicolumn{7}{c|}{$4.14$} & \multicolumn{7}{c|}{$3.14$} \\
  \cline{2-16}
  & \multicolumn{15}{|c|}{$8.33$}  \\
  \hline
 \end{tabular}
   \end{center}
\caption{Runtime for the video segmentation and its components in seconds per frame. }
\label{table:runtime_video_segmentation_total}
\end{table}

%
%
\section{Conclusions}
We have presented a video object segmentation algorithm that combines object detection with bottom-up motion and appearance cues. 
The detection makes the segmentation robust against a variety of challenges in pure bottom-up methods and provides a class label for each object instance.
In cases where the detector is not available, because the object class is unknown, the method falls back to a bottom-up approach and can still perform very well. 
Our evaluation on four video segmentation datasets showed that we achieve state-of-the-art performance except for the FBMS dataset due to the missing annotation of static objects. 
\section*{Acknowledgements}
This work was funded by the ERC Starting Grant VideoLearn.
\clearpage

\bibliographystyle{abbrvdin}
\bibliography{vidobjseg.bib}

\begin{thebibliography}{10}


\providecommand{\url}[1]{\texttt{#1}}
\expandafter\ifx\csname urlstyle\endcsname\relax
  \providecommand{\doi}[1]{doi: #1}\else
  \providecommand{\doi}{doi: \begingroup \urlstyle{rm}\Url}\fi

\bibitem[1]{Achanta2012}
\textsc{Achanta}, R.  ; \textsc{Shaji}, A.  ; \textsc{Smith}, K.  ;
  \textsc{Lucchi}, A.  ; \textsc{Fua}, P.   ; \textsc{Susstrunk}, S. :
\newblock SLIC Superpixels Compared to State-of-the-Art Superpixel Methods.
\newblock {In: }\emph{IEEE Trans. Pattern Anal. Mach. Intell.} 34 (2012), Nov.,
  Nr. 11

\bibitem[2]{Andriluka08}
\textsc{Andriluka}, M.  ; \textsc{Roth}, S.   ; \textsc{Schiele}, B. :
\newblock People-Tracking-by-Detection and People-Detection-by-Tracking.
\newblock {In: }\emph{IEEE Conference on Computer Vision and Pattern
  Recognition (CVPR)}, 2008

\bibitem[3]{Badrinarayanan2013}
\textsc{Badrinarayanan}, V.  ; \textsc{Budvytis}, I.   ; \textsc{Cipolla}, R. :
\newblock Mixture of Trees Probabilistic Graphical Model for Video
  Segmentation.
\newblock {In: }\emph{International Journal of Computer Vision}  (2013)

\bibitem[4]{Breitenstein09}
\textsc{Breitenstein}, M.~D. ; \textsc{Reichlin}, F.  ; \textsc{Leibe}, B.  ;
  \textsc{Koller-Meier}, E.   ; \textsc{Gool}, L.~V.:
\newblock Robust Tracking-by-Detection using a Detector Confidence Particle
  Filter.
\newblock {In: }\emph{IEEE International Conference on Computer Vision (ICCV)},
  2009

\bibitem[5]{Endres10}
\textsc{Endres}, I.  ; \textsc{Hoiem}, D. :
\newblock Category Independent Object Proposals.
\newblock {In: }\emph{European Conf. on Computer Vision (ECCV)}, 2010, S.
  575--588

\bibitem[6]{rcnn}
\textsc{Girshick}, R.  ; \textsc{Donahue}, J.  ; \textsc{Darrell}, T.   ;
  \textsc{Malik}, J. :
\newblock Rich feature hierarchies for accurate object detection and semantic
  segmentation.
\newblock {In: }\emph{IEEE Conference on Computer Vision and Pattern
  Recognition (CVPR)}, 2014

\bibitem[7]{Hartmann2012}
\textsc{Hartmann}, G.  ; \textsc{Grundmann}, M.  ; \textsc{Hoffman}, J.  ;
  \textsc{Tsai}, D.  ; \textsc{Kwatra}, V.  ; \textsc{Madani}, O.  ;
  \textsc{Vijayanarasimhan}, S.  ; \textsc{Essa}, I.  ; \textsc{Rehg}, J.   ;
  \textsc{Sukthankar}, R. :
\newblock Weakly Supervised Learning of Object Segmentations from Web-scale
  Video.
\newblock {In: }\emph{European Conf. on Computer Vision (ECCV)}, 2012

\bibitem[8]{Hua14}
\textsc{Hua}, Y.  ; \textsc{Alahari}, K.   ; \textsc{Schmid}, C. :
\newblock Occlusion and motion reasoning for long-term tracking.
\newblock {In: }\emph{European Conf. on Computer Vision (ECCV)}, 2014

\bibitem[9]{Jain2014}
\textsc{Jain}, S.~D. ; \textsc{Grauman}, K. :
\newblock Supervoxel-Consistent Foreground Propagation in Video.
\newblock {In: }\emph{European Conf. on Computer Vision (ECCV)}, 2014, S.
  656--671

\bibitem[10]{Kalal12}
\textsc{Kalal}, Z.  ; \textsc{Mikolajczyk}, K.   ; \textsc{Matas}, J. :
\newblock Tracking-Learning-Detection.
\newblock {In: }\emph{IEEE Trans. Pattern Anal. Mach. Intell.}  (2012), Jul.

\bibitem[11]{Keuper2015}
\textsc{Keuper}, M.  ; \textsc{Andres}, B.   ; \textsc{Brox}, T. :
\newblock Motion Trajectory Segmentation via Minimum Cost Multicuts.
\newblock {In: }\emph{IEEE International Conference on Computer Vision (ICCV)},
  2015

\bibitem[12]{Komodakis2007}
\textsc{Komodakis}, N.  ; \textsc{Tziritas}, G. :
\newblock Approximate Labeling via Graph Cuts Based on Linear Programming.
\newblock {In: }\emph{IEEE Trans. Pattern Anal. Mach. Intell.} 29 (2007), Aug.,
  Nr. 8

\bibitem[13]{Lee2011}
\textsc{Lee}, Y.~J. ; \textsc{Kim}, J.   ; \textsc{Grauman}, K. :
\newblock Key-segments for video object segmentation.
\newblock {In: }\emph{IEEE International Conference on Computer Vision (ICCV)}

\bibitem[14]{Li2013}
\textsc{Li}, F.  ; \textsc{Kim}, T.  ; \textsc{Humayun}, A.  ; \textsc{Tsai},
  D.   ; \textsc{Rehg}, J.~M.:
\newblock Video Segmentation by Tracking Many Figure-Ground Segments.
\newblock {In: }\emph{IEEE International Conference on Computer Vision (ICCV)},
  2013

\bibitem[15]{Long2015}
\textsc{Long}, J.  ; \textsc{Shelhamer}, E.   ; \textsc{Darrell}, T. :
\newblock Fully Convolutional Networks for Semantic Segmentation.
\newblock {In: }\emph{CoRR} abs/1411.4038 (2014)

\bibitem[16]{Naveen}
\textsc{Nagaraja}, N.  ; \textsc{Schmidt}, F.   ; \textsc{Brox}, T. :
\newblock Video Segmentation with Just a Few Strokes.
\newblock {In: }\emph{IEEE International Conference on Computer Vision (ICCV)},
  2015

\bibitem[17]{Ochs14}
\textsc{Ochs}, P.  ; \textsc{Malik}, J.   ; \textsc{Brox}, T. :
\newblock Segmentation of moving objects by long term video analysis.
\newblock {In: }\emph{IEEE Trans. Pattern Anal. Mach. Intell.}

\bibitem[18]{FVS}
\textsc{Papazoglou}, A.  ; \textsc{Ferrari}, V. :
\newblock Fast Object Segmentation in Unconstrained Video.
\newblock {In: }\emph{IEEE International Conference on Computer Vision (ICCV)}

\bibitem[19]{Prest13}
\textsc{Prest}, A.  ; \textsc{Ferrari}, V.   ; \textsc{Schmid}, C. :
\newblock {In: }\emph{IEEE Trans. Pattern Anal. Mach. Intell.}

\bibitem[20]{Prest12}
\textsc{Prest}, A.  ; \textsc{Leistner}, C.  ; \textsc{Civera}, J.  ;
  \textsc{Schmid}, C.   ; \textsc{Ferrari}, V. :
\newblock Learning object class detectors from weakly annotated video.
\newblock {In: }\emph{IEEE Conference on Computer Vision and Pattern
  Recognition (CVPR)}

\bibitem[21]{Rother04}
\textsc{Rother}, C.  ; \textsc{Kolmogorov}, V.   ; \textsc{Blake}, A. :
\newblock "GrabCut": Interactive Foreground Extraction Using Iterated Graph
  Cuts.
\newblock {In: }\emph{ACM Trans. Graph.} 23 (2004), Aug., Nr. 3, S. 309--314.
  --
\newblock ISSN 0730--0301

\bibitem[22]{seguin2016}
\textsc{Seguin}, G.  ; \textsc{Bojanowski}, P.  ; \textsc{Lajugie}, R.   ;
  \textsc{Laptev}, I. :
\newblock \emph{{Instance-level video segmentation from object tracks}}

\bibitem[23]{Tang13}
\textsc{Tang}, K.  ; \textsc{Sukthankar}, R.  ; \textsc{Yagnik}, J.   ;
  \textsc{Fei-Fei}, L. :
\newblock Discriminative Segment Annotation in Weakly Labeled Video.
\newblock {In: }\emph{IEEE Conference on Computer Vision and Pattern
  Recognition (CVPR)}, 2013

\bibitem[24]{Vijayanarasimhan2012}
\textsc{Vijayanarasimhan}, S.  ; \textsc{Grauman}, K. :
\newblock Active Frame Selection for Label Propagation in Videos.
\newblock {In: }\emph{European Conf. on Computer Vision (ECCV)}, 2012

\bibitem[25]{Wang2016}
\textsc{Wang}, H.  ; \textsc{Wang}, T. :
\newblock Primary object discovery and segmentation in videos via graph-based
  transductive inference.
\newblock {In: }\emph{Computer Vision and Image Understanding} 143 (2016)

\bibitem[26]{Weinzaepfel2015}
\textsc{Weinzaepfel}, P.  ; \textsc{Harchaoui}, Z.   ; \textsc{Schmid}, C. :
\newblock Learning to Track for Spatio-Temporal Action Localization.
\newblock {In: }\emph{IEEE International Conference on Computer Vision (ICCV)},
  2015

\bibitem[27]{yang2016}
\textsc{Yang}, J.  ; \textsc{Price}, B.~L. ; \textsc{Shen}, X.  ; \textsc{Lin},
  Z.~L.  ; \textsc{Yuan}, J. :
\newblock Fast Appearance Modeling for Automatic Primary Video Object
  Segmentation.
\newblock {In: }\emph{IEEE Trans. on Image Processing}  (2016), Feb

\bibitem[28]{Yang2015}
\textsc{Yang}, Y.  ; \textsc{Sundaramoorthi}, G.   ; \textsc{Soatto}, S. :
\newblock Self-Occlusions and Disocclusions in Causal Video Object
  Segmentation.
\newblock {In: }\emph{IEEE International Conference on Computer Vision (ICCV)},
  2015

\bibitem[29]{Dong14}
\textsc{Zhang}, D.  ; \textsc{Javed}, O.   ; \textsc{Shah}, M. :
\newblock Video Object Segmentation through Spatially Accurate and Temporally
  Dense Extraction of Primary Object Regions.
\newblock {In: }\emph{IEEE Conference on Computer Vision and Pattern
  Recognition (CVPR)} 0 (2013), S. 628--635. --
\newblock ISSN 1063--6919

\bibitem[30]{Zhang15}
\textsc{Zhang}, Y.  ; \textsc{Chen}, X.  ; \textsc{Li}, J.  ; \textsc{Wang}, C.
    ; \textsc{Xia}, C. :
\newblock Semantic Object Segmentation via Detection in Weakly Labeled Video.
\newblock {In: }\emph{IEEE Conference on Computer Vision and Pattern
  Recognition (CVPR)}

\bibitem[31]{Zitnick14}
\textsc{Zitnick}, C.~L. ; \textsc{Doll\'ar}, P. :
\newblock Edge Boxes: Locating Object Proposals from Edges.
\newblock {In: }\emph{European Conf. on Computer Vision (ECCV)}

\end{thebibliography}
\end{document}